\documentclass[letterpaper, 10 pt, conference]{ieeeconf}  

\IEEEoverridecommandlockouts                              

\usepackage[utf8]{inputenc}
\usepackage[T1]{fontenc}
\usepackage[svgnames]{xcolor}

\usepackage{graphicx}
\usepackage{epstopdf}
\usepackage{subcaption}
\usepackage{hyperref}
\definecolor{darkgreen}{rgb}{0,0.5,0}
\definecolor{frenchblue}{rgb}{0.0, 0.45, 0.73}
\hypersetup{
    unicode=false,          
    colorlinks=true,        
    linkcolor=darkgreen,          
    citecolor=darkgreen,        
    filecolor=magenta,      
    urlcolor=blue           
}

\usepackage{mathtools}
\DeclarePairedDelimiterX{\norm}[1]{\lVert}{\rVert}{#1}
\DeclarePairedDelimiterX{\abs}[1]{\lvert}{\rvert}{#1}

\usepackage{amsmath}
\usepackage{amsfonts}
\usepackage{amssymb}
\usepackage[capitalize, nameinlink]{cleveref}
\usepackage{bm}
\usepackage{bbm}
\usepackage{mathdots}
\usepackage{multicol}
\usepackage{multirow}
\let\labelindent\relax
\usepackage{enumitem}
\usepackage{booktabs}
\usepackage{algorithm}
\usepackage{algpseudocode}
\usepackage[backgroundcolor=gray!20, linecolor=gray, textsize=scriptsize]{todonotes}
\usepackage{graphicx}
\usepackage{forest}
\usepackage{tikz}
\usepackage{authblk}
\usetikzlibrary{calc, graphs, graphs.standard, shapes, arrows, positioning, decorations.pathreplacing, decorations.markings, decorations.pathmorphing, fit, matrix, patterns, shapes.misc}
\usepackage{standalone}
\usepackage{makecell}

\renewcommand{\arraystretch}{1.2}

\newcommand{\green}[1]{{\color{darkgreen}{#1}}}
\newcommand{\red}[1]{{\color{red}{#1}}}

\crefname{claim}{Claim}{Claims}

\title{\LARGE \bf
Depth Estimation from Monocular Images and Sparse Radar Data}

\author[1]{Juan-Ting Lin}
\author[1]{Dengxin Dai}
\author[1]{Luc Van Gool}
\affil[1]{Computer Vision Lab, ETH Zurich, Switzerland}
\affil[ ]{\tt\small julin@student.ethz.ch, \tt\small \{dai, vangool\}@vision.ee.ethz.ch}

\begin{document}

\maketitle
\thispagestyle{empty}
\pagestyle{empty}

\begin{abstract}

In this paper, we explore the possibility of achieving a more accurate depth estimation by fusing monocular images and Radar points using a deep neural network. 
We give a comprehensive study of the fusion between RGB images and Radar measurements from different aspects and proposed a working solution based on the observations. 
We find that the noise existing in Radar measurements is one of the main key reasons that prevents one from applying the existing fusion methods developed for LiDAR data and images to the new fusion problem between Radar data and images. The experiments are conducted on the nuScenes dataset, which is one of the first datasets which features Camera, Radar, and LiDAR recordings in diverse scenes and weather conditions. 
Extensive experiments demonstrate that our method outperforms existing fusion methods. We also provide detailed ablation studies to show the effectiveness of each component in our method. Our code will be released in the following link: \href{https://github.com/brade31919/radar_depth}{https://github.com/brade31919/radar\_depth}.

\end{abstract}

\section{INTRODUCTION}
\label{sec:intro}

Dense and robust depth estimation is an important component in self-driving system and unmanned aerial vehicles. While existing structure-light-based depth sensor or stereo camera can provide dense depth in indoor environments~\cite{nyudepthv2}, the reliability of these sensors degrade a lot in outdoor applications. 
As a result, lots of research works focus on obtaining dense depth from monocular RGB images only. 
Recently, convolutional neural network (CNN) based methods have demonstrated impressive improvements on monocular depth estimation for both indoor and outdoor scenarios~\cite{dorn, depth-cnn1, fcrn,reccurent_depth_2020}. 
However, there is still a gap between the accuracy and reliability of these methods and what the real-world applications need.

Apart from estimating depth from monocular camera, to improve the robustness of the system, some methods also take other sensor modalities into consideration. Within these sensors, LiDAR is the most commonly used one. Many works have been conducted on dense depth estimation from RGB images and sparse LiDAR scans~\cite{deeplidar, sparse2dense}. In addition to depth estimation and completion tasks, different RGB + LiDAR fusion techniques are also extensively used in tasks such as 3D object detection~\cite{uber-conti-2, uber-conti-3}. 
Although LiDAR provides more accurate depth measurements in outdoor scenario, high-end LiDAR sensors are still far from affordable for many applications. 

Compared with LiDAR, Radar is an automotive-grade sensor that has been used for decades on vehicles, but has not attracted lots of attention in self-driving research based on deep learning. 
One reason might be that Radar measurements are not included in most of the dominant self-driving datasets~\cite{kitti}. 
Compared with LiDAR, Radar sensors offer longer sensing range (200m $\sim$ 300m), more attributes including velocities, dynamic states, and measurement uncertainties. Most importantly, the costs of these sensors are much lower than LiDAR.
However, Radar measurements are typically sparser, noisier, and have a more limited vertical field of view.

This work is to study the challenges of using Radar data for dense depth estimation and to propose a novel method for that aim. Given recently released nuScenes dataset \cite{nuscenes2019} consisting of RGB, LiDAR, and Radar measurements, we are able to conduct experiments on cross-modality sensor fusion between RGB camera and Radar. Through our experiments, we demonstrated that:
1) Existing RGB + LiDAR fusion methods can not be applied directly to RGB + Radar fusion task; and 2) with proper fusion strategies and a novel denoising operation, our proposed network is able to improve the performance of depth estimation by a good margin by using Radar measurements. According to our survey, our work is the first one that brings Radar sensors into dense depth estimation tasks.

The contributions of this work include: 1) a detailed study on the challenges of using Radar data for dense depth estimation; and 2) a novel and carefully motivated network architecture for depth estimation with monocular images and sparse Radar data.
\section{RELATED WORKS}
\label{sec:rw}



\begin{figure*}[t]
    \centering
    \includegraphics[width=0.92\textwidth]{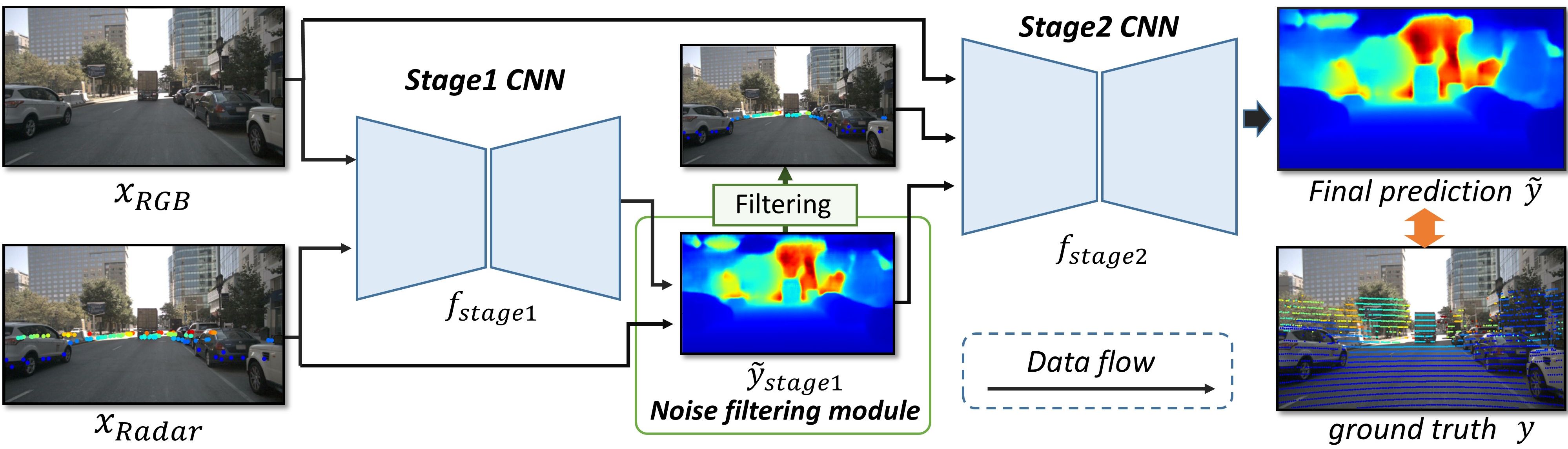}
    \caption{The full pipeline of our method. Each CNN stage is a full late fusion model described in \cref{subsec:cnn}.}
    \vspace{-4mm}
    \label{fig:full_model}
\end{figure*}

\vspace{+2mm}
\textbf{RGB-based Depth Estimation.}
Depth estimation from monocular or stereo camera is a popular research topic in both computer vision and robotics. Early works used either geometry-based algorithms on stereo images \cite{early_stereo1, early_stereo2} or handcrafted features on single images \cite{make3d, early_mono1, early_mono2}. 
Recent years, convolutional neural networks (CNN) have demonstrated their ability in image understanding~\cite{vgg, resnet}, dense predictions~\cite{fcn, mask-rcnn}, etc. given large scale datasets~\cite{kitti,cityscapes, mscoco}. Therefore, lots of research works of monocular depth estimation~\cite{eigen, depth-cnn1, depth-cnn2, fcrn, dorn} are conducted. 
In general, most of them used the encoder-decoder architectures~\cite{fcrn, dorn}. Xie et al.~\cite{depth-skip} further introduced skip-connection strategy which is a frequently used technique to multi-level features in dense prediction tasks. On the other hand, Huang et al.~\cite{dorn} achieved state-of-the-art performance by introducing space increasing discretization (SID) and ordinal regression. 
In some semi- / self-supervised formulations, photometric reconstruction error instead of L1/L2 loss is used~\cite{sfmlearner, unsupervised-photo}, and smoothness constraint is further imposed~\cite{sfmlearner, semi-smooth, monodepth2, edge-aware-smoothness} to enhance local consistency. Patil et al. \cite{reccurent_depth_2020} proposed a recurrent network architecture to exploit the long-range spatiotemporal structures across video frames to yield more accurate depth maps.  
While good performance has been obtained with only RGB images, the methods still have difficulty in generalizing to new scenarios and challenging weather and lighting conditions. This motivates the existing line of work that fuses camera data with LiDAR data and our work that fuses camera data with Radar data which is cheaper to obtain. 



\textbf{Depth Estimation with Camera and LiDAR Data.}
While monocular depth estimation task attracts lots of attention, achieving more reliable and accurate predictions using multi-modality information is also a popular topic. 
Existing works either take the whole set of LiDAR points~\cite{deeplidar, completion1} (known as \textit{depth completion}), or the downsampled set as model inputs~\cite{sparse2dense, self-sparse2dense} (known as \textit{depth prediction/estimation}).
Ma et al.~\cite{sparse2dense} first projected LiDAR points to 2D sparse depth map and then perform so called early fusion by direct concatenation with RGB images along channel, or concatenating feature maps after one shallow convolution block~\cite{self-sparse2dense}. 
Jaritz et al.~\cite{completion1} used a late fusion method to combine features from different modalities and improved the overall performance through multitask learning. 
Qiu et al. \cite{deeplidar} proposed to predict dense depth map by combining predictions from RGB and surface normal pathways, where surface normal is treated as an intermediate representation. Moreover, confidence maps are predicted to down-weight mixed measurements from LiDAR caused by the displacement between camera and LiDAR. 
In this work, our main focus is sensor fusion. Thus, We use the widely adopted encoder-decoder architecture and focus on the necessary extensions in order to effectively use Radar data instead. 

\textbf{Post-processing and Refinement Methods.}
Apart from treating sparse point clouds as inputs to the model, some methods also tried to directly refine the dense predictions of the trained models.
Wang et al.~\cite{pnp} proposed a simple add-on module that can improve the prediction of depth estimation model using similar methods used by white box adversarial attack~\cite{adversarial}. 
Since the refinement is done using iterative re-inference, no re-training is required. This method can be integrated into most deep learning based methods.
Cheng et al.~\cite{cspn} learned an affinity matrix from data to refine the outputs of their CNN model. The recurrent refinement operation can also be extended to depth completion tasks.

\textbf{Fusion of Images and Radar Data}. 
There are already works that fuse RGB images and Radar, given the fact that they are very much complementary. This line of work mainly focus on object detection and tracking. For instance, Chadwick et al.~\cite{Chadwick2019DistantVD} fused Radar data and images to detect small objects at a large distance. In \cite{nobis19crfnet} and \cite{john2019so}, the authors enhance current 2D object detection networks by fusing camera data and projected sparse radar data in the network layers, while \cite{john2019so} also performs free space semantic segmentation jointly. Both methods learn at which level the fusion
of the sensor data is more beneficial for the task. In addition to the nuScenes dataset~\cite{nuscenes2019}, there are also other 
datasets proposed for object detection with Radar data such as \cite{radar:dataset:19}.
Exemplary works on semantic segmentation with Radar point cloud have been conducted as well. For instance, in \cite{semseg:radar:points:cloud:18} the authors have studied how the challenging task can be performed and provide results on a dataset with manually labeled radar reflections. Similar to these works for object detection and semantic segmentation, our work aims to study how the challenging task of dense depth estimation with Radar data can be addressed with the popular deep neural network architectures.

In the line of increasing the robustness of depth estimation, Vasudevan et al. \cite{soundperception20} have proposed a novel method to estimate depth maps based on binaural sounds. 
\section{METHOD}
\label{sec:method}

Our whole method are divided into multiple main components.
In the following subsections, we will go through each component in details. 

\subsection{Radar data background}
\label{subsec:Radar_background}

Different from well established depth completion~\cite{deeplidar, completion1} or depth estimation tasks~\cite{dorn, sparse2dense}, there's no prior research works on RGB + Radar depth estimation task. 
Therefore, we provide a brief introduction to the task formulation and some key differences between Radar and LiDAR measurements, which will help readers to understand the motivations behind the components of our method.

\textbf{Data format.} Similar to LiDAR data, Radar measurements are recorded as sparse point clouds. 
The main difference is that, in addition to $x$, $y$, $z$, and $reflectance$, Radar data consist of additional measurements including the velocity along x and y direction, the standard deviation of location and velocity measurements, and information such as the dynamic states of the measured object (encoded as discrete numbers)\footnotemark.

\begin{figure}[t]
    \centering
    \includegraphics[width=0.48
    \textwidth]{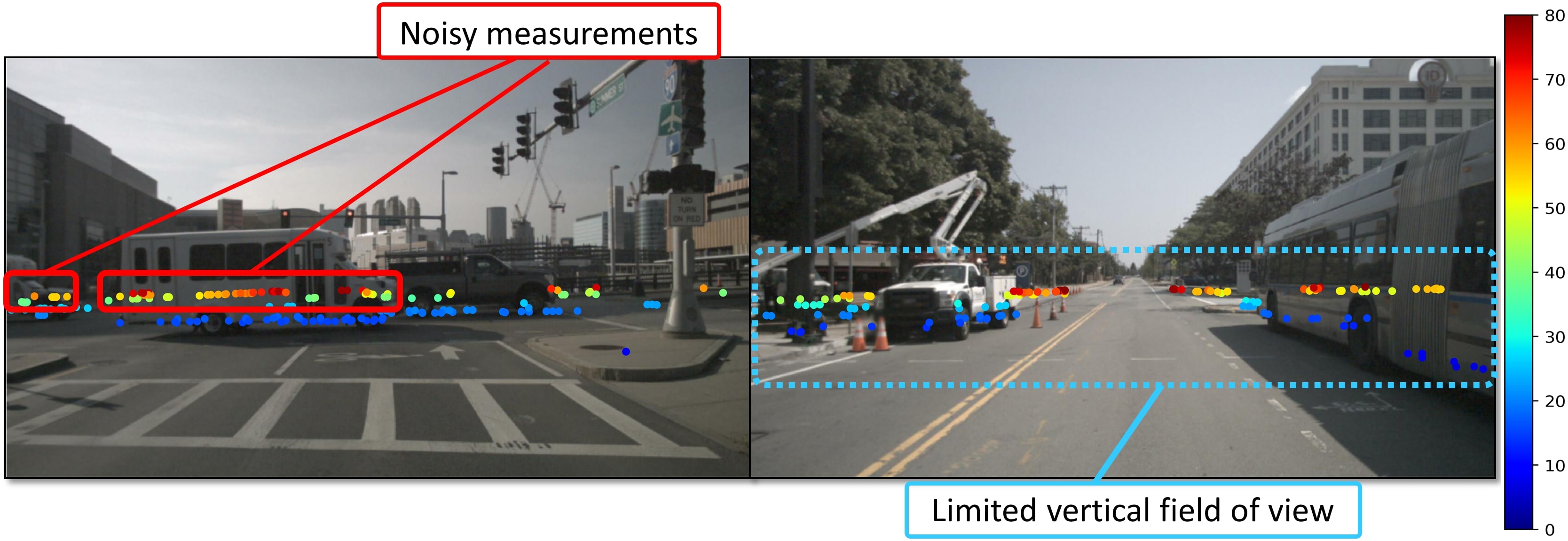}
    \caption{Limitations of Radar measurements. (Left) Noisy measurements and (Right) Limited vertical field of view.}
    \vspace{-6mm}
    \label{fig:Radar_issue}
\end{figure}

\textbf{Limitations.} While it seems that the Radar data provide more information, it also introduces the following limitations compared with LiDAR data: 
\begin{itemize}
    \item Sparseness: In nuScenes dataset~\cite{nuscenes2019}, there are more than 3000 LiDAR points after projection to the camera. However, there are less than 100 Radar points after the projection (\cref{subsec:dataset}).
    \item Limited vertical field of view: Because of the limitation of the sensor, Radar measurements mainly concentrate in the central horizontal bin (similar heights) as shown in \cref{fig:Radar_issue} (right).
    \item Noisy measurements: Due to multiple reflections (Radar multipath problem) or other reasons, we have many noisy measurements as shown in \cref{fig:Radar_issue} (left).
    \item Inconsistency with LiDAR data: Apart from noisy measurements, which are considered as outliers, the 3D points of Radar and LiDAR representing the same object can also be different. Since we typically use LiDAR measurements as ground truth~\cite{kitti}, even noise-free Radar measurements are not perfect on the evaluation metrics.
\end{itemize}

As we will show in \cref{sec:exp}, using Radar depth maps directly as the inputs of off-the-shelf RGB + LiDAR depth completion / prediction models resulted in marginal improvements.

\footnotetext{For more details, please visit \href{https://github.com/nutonomy/nuscenes-devkit/blob/master/python-sdk/nuscenes/utils/data_classes.py\#L313}{https://github.com/nutonomy/nuscenes-devkit/blob/master/python-sdk/nuscenes/utils/data\_classes.py\#L313}}

\textbf{Problem formulation.} In our RGB + Radar formulation, each data sample from the dataset contains (1) an RGB image $x_{RGB}$, (2) a set of Radar measurements $R=\{r_{n}\}^{N}_{n=1}$ from 3 nearest timestamps, and (3) a set of LiDAR measurements $L=\{l_{m}\}^{M}_{m=1}$. Radar measurements $R$ can be further projected to a single-channel 2D depth map $x_{Radar}$ using the perspective projection. Similarly, LiDAR measurements can be projected to 2D map $y$, which is treated as ground truth depth map in our experiments (\cref{sec:exp}). Our model takes both $x_{RGB}$ and $x_{Radar}$ as inputs and predicts dense 2D depth map $\Tilde{y}$ which minimizes the metric errors. Same as all the depth estimation / completion tasks, loss and metric error are computed over the pixels with ground truth measurements.

\begin{figure}[t]
    \centering
    \includegraphics[width=0.46\textwidth]{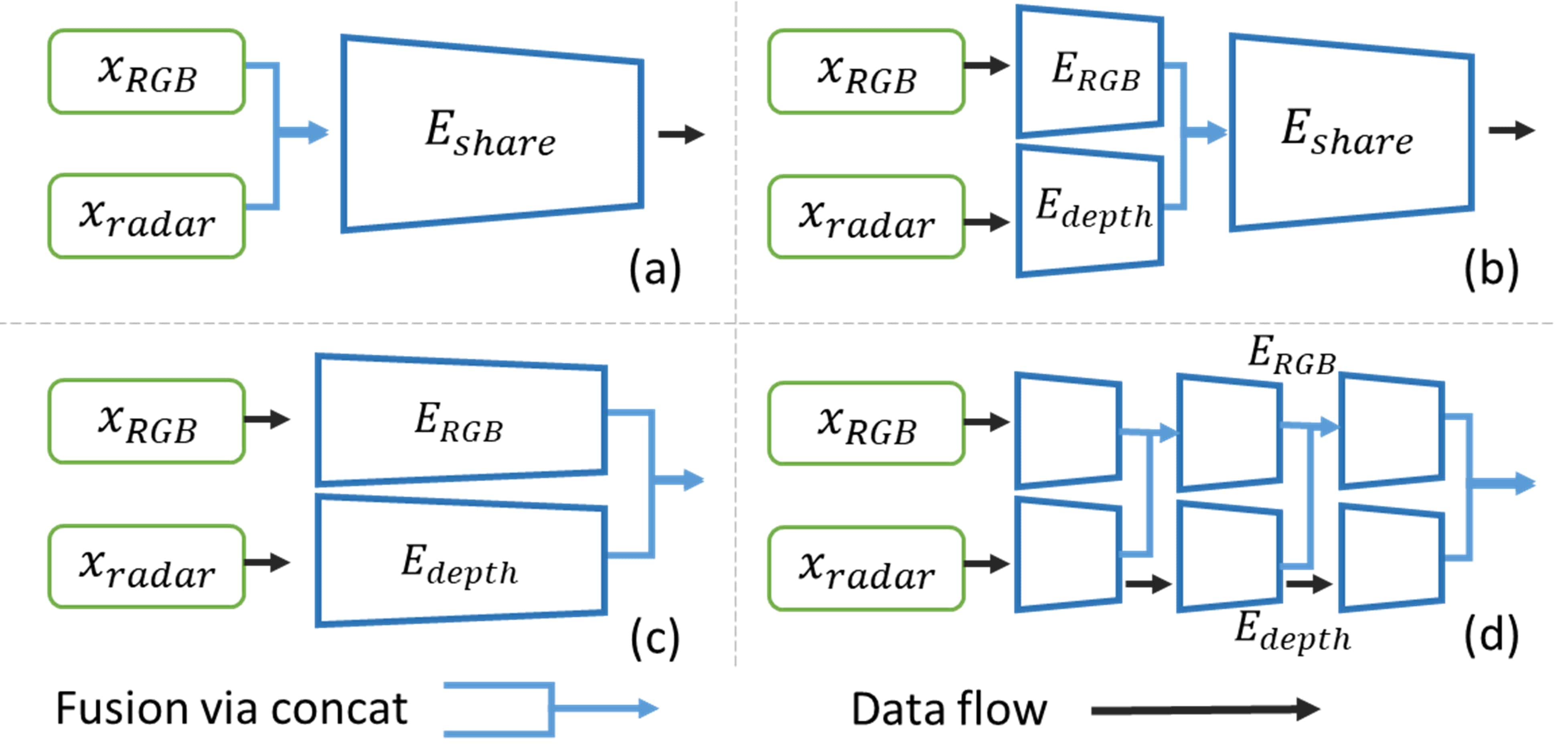}
    \caption{Fusion methods for encoder: (a) Early fusion, (b) Mid fusion, (c) Late fusion, and (d) Multi-layer fusion. Due to space limit, we didn't show details layer-wise.}
    \vspace{-4mm}
    \label{fig:fusion_illustration}
\end{figure}

\subsection{CNN architecture}
\label{subsec:cnn}

\textbf{Backbone.} Since our goal is to perform a comprehensive study on different fusion methods, we need to choose an efficient and widely-used backbone. 
Thus, we fixed our backbone to Resnet18~\cite{resnet}, and explored different fusion methods based on it as a pilot experiment. 
As illustrated in \cref{fig:fusion_illustration}, we apply different encoder fusion methods to the simple encoder-decoder architecture proposed by Ma et al.~\cite{sparse2dense} and compare their performance. 
According to the experiment (\cref{subsec:exp-pilot}), late fusion and multi-layer fusion model have comparable performance. 
Therefore, we adopt late fusion as our main encoder backbone design in the following experiments for simplicity.

\textbf{Two-Stage architecture.} As shown by the pilot experiments (\cref{subsec:exp-pilot}), a simple encoder-decoder architecture can have some improvements on RGB + Radar depth prediction task if we can remove most of the noisy measurements. 
However, we don't have the LiDAR ground truth to help us performing the filtering in real applications, and it's hard to perform outlier rejections without information on the 3D structure or objects of the scene.
Therefore, we come up with a 2-stage design to address the noisy measurement issue. 

As shown in \cref{fig:full_model}, our whole method contains two stages. The stage1 model $f_{stage1}$ takes both the RGB image $x_{RGB}$ and the Radar depth map $x_{Radar}$ as inputs and predicts a coarse depth map $\Tilde{y}_{stage1}$, which gives us a dense 3D structure of the scene:
\vspace{-2mm}
\begin{align}
    \Tilde{y}_{stage1} = f_{stage1}(x_{RGB}, x_{Radar})
\end{align}
\vspace{-4mm}

Then, we compare the Radar depth map with the coarse prediction $\Tilde{y}_{stage1}$ to reject some outliers (more details in next subsection \cref{subsec:noise_filter}) and obtain the filtered Radar depth map $\Tilde{x}_{Radar}$.
The assumption here is that although the predictions from stage1 is not perfect, they are smooth and locally consistent. 
Therefore, they are suitable to reject outlier noises produced by Radar multipath, which typically have certain margins with the correct depth values.

The stage2 model $f_{stage2}$ takes $x_{RGB}$, $\Tilde{x}_{Radar}$, and the prediction from stage1 $\Tilde{y}_{stage1}$ to predict the final result $\Tilde{y}$:
\vspace{-2mm}
\begin{align}
    \Tilde{y} = f_{stage2}(x_{RGB}, \Tilde{x}_{Radar}, \Tilde{y}_{stage1}).
\end{align}

\subsection{Noise filtering module}
\label{subsec:noise_filter}
Since Radar measurements are not exactly consistent with the LiDAR measurements as we mentioned in \cref{subsec:Radar_background}, we need to have some tolerances in the filtering process. Otherwise, we will end up discarding all the measurements in the set $R$. 

Instead of setting a fixed distance tolerance threshold $\tau$, we empirically found that an adaptive threshold gives us better results. We design the threshold to be a function of depth value $\tau(d)$: We have larger tolerance for large depth values, which is similar to the space-increasing disrectization (SID) from Huan et al.~\cite{dorn}:
\begin{align}
    \tau(d) = \exp{(\frac{d*\log(\frac{\beta}{\alpha})}{K} + \log(\alpha))},
\end{align}
here we heuristically set $\alpha=5$ and $\beta=18$.

Let $P$ denote the set pixel coordinates $(u, v)$ of the Radar measurements projected by the perspective projection function $proj(.)$: $P=\{p_{n}\}^{N}_{n=1} = \{proj(r_{n})\}^{N}_{n=1}$. The noise filtering module will keep the point $p_{n}$ if it satisfies the follow constraint:
\begin{align}
    \abs{x_{Radar}(p_{n}) - \Tilde{y}_{stage1}(p_{n})} \leq \tau(p_{n}), \, \textnormal{for $p_{n} \in P$ }
\end{align}

\subsection{Loss Functions}
\label{subsec:lossfunction}

\textbf{Loss functions.} By design, each component of our model is differentiable. Thus, our whole model is end-to-end trainable. Following the setting from~\cite{sparse2dense}, we apply L1 loss to both the predictions of stage1 ($\Tilde{y}_{stage1}$) and stage2 ($\Tilde{y}$). Considering that the main purpose of $\Tilde{y}_{stage1}$ is to filter outlier noises in $x_{Radar}$, we further add edge-aware  smoothness constraint~\cite{monodepth2, edge-aware-smoothness} to it. To effectively balance multiple loss terms, we follow the method proposed by Kendall et al.~\cite{multi_uncertainty}:
\vspace{-1mm}
\begin{align}
    \label{eq:total}
    \mathcal{L}_{total} = &e^{-w_{1}} * (L1(\Tilde{y}_{stage1}, y) + 10^{-3}*\mathcal{L}_{smooth}) + \\
    &e^{-w_{2}}*L1(\Tilde{y}, y) + \sum_{i} w_{i},
\end{align}
\vspace{-1mm}
where $w_{1}$ and $w_{2}$ are optimized variables, and $\mathcal{L}_{smooth}$ is defined as:
\begin{align}
    \mathcal{L}_{smooth} = &\abs{\nabla_{u} (\Tilde{y}_{stage1})}e^{-\abs{\nabla_{u}(x_{RGB})}} + \\
    &\abs{\nabla_{v} (\Tilde{y}_{stage1})}e^{-\abs{\nabla_{v}(x_{RGB})}}
\end{align}
$\nabla_{u}$ and $\nabla_{v}$ denote the gradient along 2D height and width directions separately.

\subsection{Implementation details}
\label{subsec:implementation}

Unless stated otherwise, all the models are trained using a batch size of 16 and the SGD optimizer with a learning rate of 0.001 and a momentum of 0.9 for 20 epochs. The learning rate is multiplied by 0.1 after every 5 epochs. All the models we used in experiment section are implemented in PyTorch~\cite{pytorch}. The experiments are conducted on desktop computers / clusters with Nvidia GTX1080Ti and TeslaV100 GPUs.

All of our encoder network architectures (\cref{subsec:cnn}) are modified from the standard Resnet18~\cite{resnet}. 
For early fusion, we simply modified the input channels to 4 and randomly initialized the weights (weights of other layers were initialized from pre-trained models on Imagenet dataset~\cite{imagenet}). 
For mid, late, and multi-layer fusion, the depth branch has a similar architecture as the RGB branch. The only difference is that we change the number of channels to 1/4 of the original one. 
Fusion operations happened only on feature maps with same spatial resolution (width and height). Regarding the decoder part, we kept the setting from~\cite{sparse2dense, self-sparse2dense, fcrn} by using the UpProj module as our upsampling operation.

\section{EXPERIMENT}
\label{sec:exp}

\subsection{Dataset}
\label{subsec:dataset}
Since we conducted all our experiments on the newly released nuScenes dataset \cite{nuscenes2019}. There's no previous works to reference for details such as train/val splits, point cloud projection, preprocessing steps, etc. 
Therefore, we will provide some background information of the dataset itself, how we setup the evaluation process, and also some visualizations.

The nuScenes dataset \cite{nuscenes2019} is the first dataset providing full autonomous vehicle sensor suite including 6 cameras, 5 Radars and 1 LiDAR. 
All types of sensor have roughly covered 360 degree field of view. Extrinsics relative to car ego frame are provided to project point clouds to different sensor coordinates. 
More details of the sensor setup, calibration, and synchronization can be found from the official site: \href{https://www.nuscenes.org/data-collection}{https://www.nuscenes.org/data-collection}.

The dataset is organized by scenes. Each scene is a 20s long driving sequence, and roughly 40 samples are selected from single sequence to have synchronized sensor recordings. 
The whole dataset contains 1000 scenes, but only the annotations of trainval split are publicly available (850 scenes).
Therefore, we used the available 850 scenes and split them into 765 scenes for training and 85 scenes for evaluation. 
As a result, we have 30750 samples in the training set and 3399 samples in the evaluation set.

In each sample, we have 6 images and 5 Radar sweeps with different orientations. 1 LiDAR sweep with omnidirectional view. For the depth estimation task, we didn't use all orientations\footnotemark. 
In our experiment setting, we only used the "front" and "back" views of each sample for simplicity. 
Therefore, we have totally $30750 \times 2=61500$ training data, and $3399 \times 2=6798$ evaluation data, which is comparable to the scale of existing outdoor depth estimation/completion dataset \cite{kitti}. 
Each datum contains:

\footnotetext{Note that the original nuscenes dataset only offered detection and tracking benchmark. 
The depth estimation dataset is our own re-organization}

\begin{figure}[t]
\centering
\includegraphics[width=0.48\textwidth]{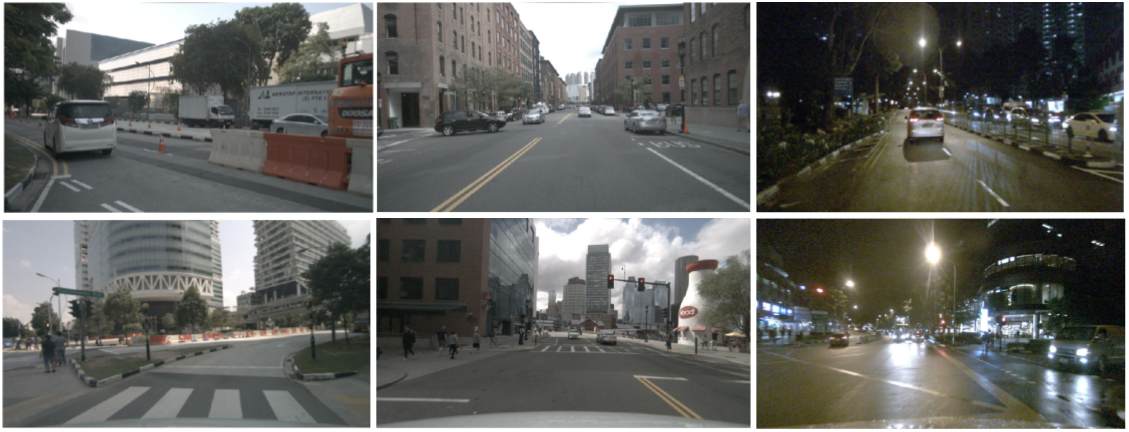} 
\vspace{-3mm}
\caption{Sample views. Top row are front views and bottom row are back views. The last column are the night views.}
\vspace{-1mm}
\label{fig:image-examples}
\end{figure}

\begin{itemize}
    \item RGB image $x_{RGB}$ with $450\times800$ resolution which is downscaled from original $900\times1600$ resolution. Check \cref{fig:image-examples} for some examples in the dataset.
    \item LiDAR depth map $y$ with $450\times800$ resolution (Around 3000-5000 points / image).
    \item Radar depth map $x_{Radar}$ with $450\times800$ by projecting all the point clouds from Radar sensors having overlapping field of view with the target camera. In our experiment setting, we will accumulate Radar points from 3 nearest timestamps to increase the point counts. (Around 40-100 points / image) 
\end{itemize}

\begin{figure}[t]
\centering
\includegraphics[width=0.48\textwidth]{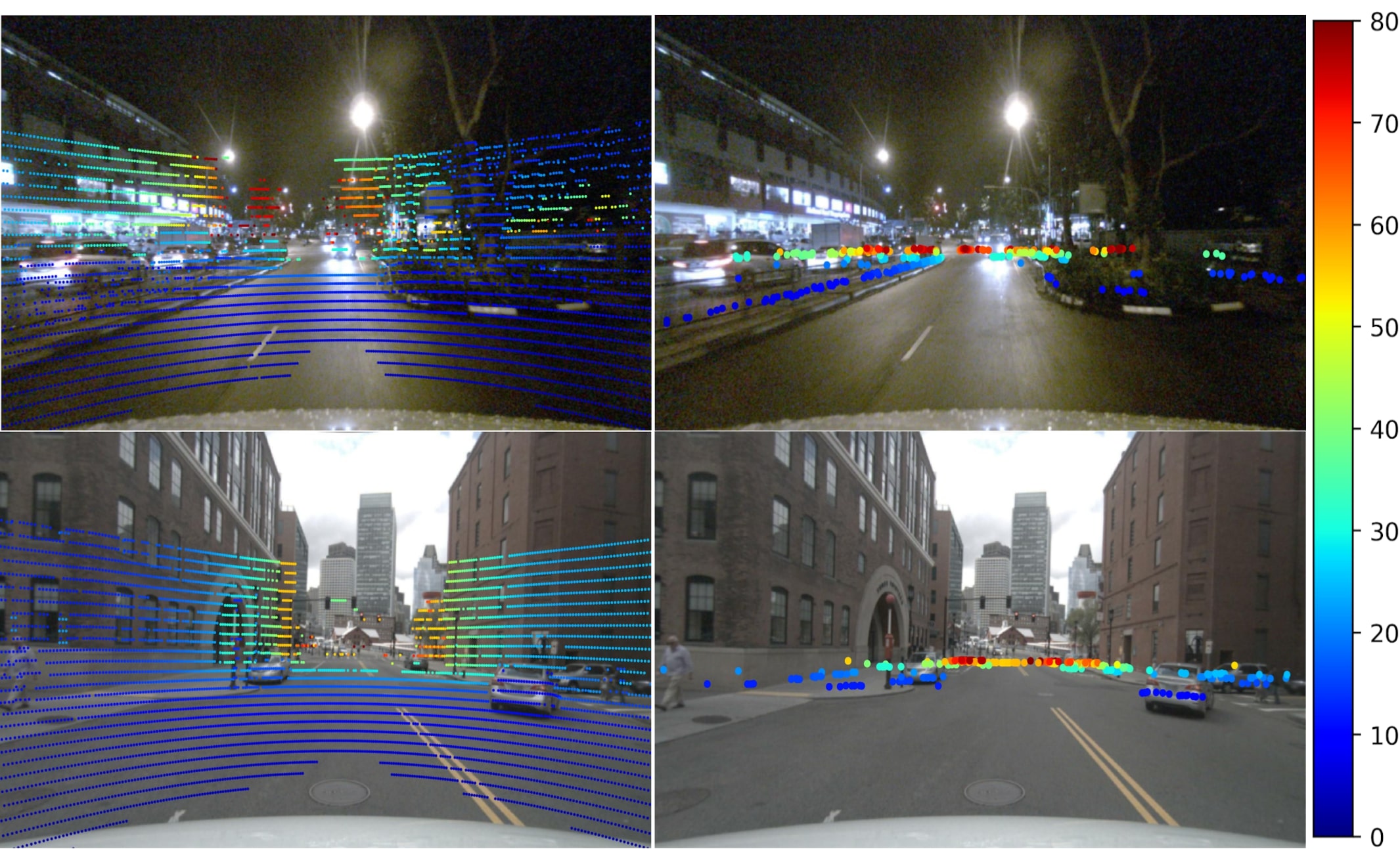} 
\caption{nuScenes\cite{nuscenes2019} sensor data visualization. The left column contains LiDAR results and right column contains Radar results. We use point size larger than 1 pixel for better visualization.}
\vspace{-4mm}
\label{fig:lidar_Radar-examples}
\end{figure}

\cref{fig:lidar_Radar-examples} shows the rendered point clouds on images.

\subsection{Evaluation metrics}
\label{subsec:metrics}

Since the predictions are dense depth maps, we use the mostly adopted metrics from depth estimation / completion tasks~\cite{eigen, sparse2dense}. Here we used $T$ to denotes the set of 2D pixels with ground truth LiDAR depth values, $y$ and $\Tilde{y}$ denote the ground truth and predicted depth maps separately. $y(p_{i})$ denotes the depth value corresponds to pixel $p_{i}$ where $p_{i} \in T$:

\begin{itemize}
    \item Root Mean Square Error (RMSE): \\
    \hspace*{5mm} $\sqrt{\frac{1}{|T|}\sum_{p_{i} \in T} \norm{y(p_{i}) - \tilde{y}(p_{i})}_{2}^2}$
    \vspace{1mm}
    \item Mean Absolute Error (MAE): \\
    \hspace*{5mm} $\frac{1}{|T|} \sum_{p_{i} \in T} \abs{y(p_{i}) - \tilde{y}(p_{i})}$
    \vspace{1mm}
    \item Mean Absolute Error logscale ($\text{MAE}_{\text{log}}$): \\
    \hspace*{5mm} $\frac{1}{|T|} \sum_{p_{i} \in T} \abs{\log{y(p_{i})} - \log{\tilde{y}(p_{i})}}$
    \vspace{1mm}
    \item Mean Absolute Relative Error (REL): \\ \hspace*{5mm} $\frac{1}{|T|} \sum_{p_{i} \in T} \abs{y(p_{i}) - \tilde{y}(p_{i})}\, /\, {y(p_{i})}$
    \vspace{1mm}
    \item $\delta_{n}$ threshold: \\
    \hspace*{5mm} $\delta_{n} = \abs{\{ \tilde{y}(p_{i}):\, max(\frac{\tilde{y}(p_{i})}{y(p_{i})}, \frac{y(p_{i})}{\tilde{y}(p_{i})}) < 1.25^{n}\}} \, / \, {\abs{T}}$.
\end{itemize}

\subsection{Pilot experiments}
\label{subsec:exp-pilot}

\textbf{Encoder architecture.} We replace the Resnet18~\cite{resnet} encoder in Sparse-to-dense~\cite{sparse2dense} by the encoders using different fusion techniques mentioned in \cref{subsec:cnn} (implementation details in \cref{subsec:implementation}). According to \cref{tab:exp-fusion}, we can see that late fusion gives us the best performance over other fusion techniques. 

Therefore, we choose late fusion encoder design as the main backbone component of our CNN model in the following experiments. 
To show the improvements over monocular depth prediction, we also include RGB only model (first row) into comparison. 
We can find that we can only achieve marginal improvements if we use the raw Radar depth maps $x_{RGB}$ directly as the inputs to the standard encoder-decoder architecture (early fusion), which implies that further processing steps on noisy Radar measurements are non-trivial for better performance.

\renewcommand{\arraystretch}{1.5}
\begin{table}[t]
    \centering
    \begin{tabular}{c|c|cccc}
    \toprule 
    \makecell{Fusion \\ method} & \makecell{Depth \\inputs} & $\delta_{1}\uparrow$ & RMSE $\downarrow$ & MAE $\downarrow$ & REL $\downarrow$ \\
    \midrule
    \midrule
    RGB only & None & 0.862 & 5.613 & 2.504 & 0.126\\
    \cline{1-6}
    Early fusion & \multirow{4}{*}{Radar} & 0.876 & 5.628 & 2.374 & 0.115\\
    Mid fusion & & 0.875 & \textbf{5.243} & 2.293 & 0.121\\
    Late fusion & &\textbf{0.884} & 5.409 & \textbf{2.27} & \textbf{0.112}\\
    Multi-layer fusion & & 0.876 & 5.623 & 2.371 & 0.116\\
    \bottomrule
    \end{tabular}
    \caption{Comparison between different fusion methods.}
    \label{tab:exp-fusion}
\end{table}
\renewcommand{\arraystretch}{1.0}
\begin{table}[t]
    \centering
    \begin{tabular}{c|c|cccc}
    \toprule 
    Methods & \makecell{Depth input \\ pattern} & $\delta_{1}\small{\uparrow}$ & RMSE$\small{\downarrow}$ & MAE$\small{\downarrow}$  & REL$\small{\downarrow}$  \\
    \midrule
    \midrule
    RGB only & \textit{None} & 0.862 & 5.613 & 2.504 & 0.126 \\
    \midrule
    Radar & \textit{Radar} & 0.876 & 5.628 & 2.374 & 0.115 \\
    \cline{1-2}
    \makecell{Radar \\(filtered)} & \makecell{\emph{Radar-gt-}\\\emph{filtered}} & 0.877 & 5.630 & 2.367 & 0.115 \\
    \cline{1-2}
    \makecell{LiDAR \\(Radar pattern)} & \textit{lidar-sampled} & 0.887 & 5.364 & 2.212 & 0.109 \\
    \cline{1-2}
    \makecell{LiDAR \\(uniform)} & \textit{lidar-uniform} & 0.898 & 5.447 & 2.084 & 0.097 \\
    \bottomrule
    \end{tabular}
    \caption{Comparison between different depth input patterns.}
    \vspace{-4mm}
    \label{tab:exp-pattern}
\end{table}

\renewcommand{\arraystretch}{1.2}
\begin{table*}[t!]
    \centering
    \begin{tabular}{c|c|ccccccc}
    \toprule 
    Methods & Depth input pattern & $\delta_{1} \uparrow$ & $\delta_{2} \uparrow$ & $\delta_{3} \uparrow$ & RMSE $\downarrow$ & MAE $\downarrow$ & REL $\downarrow$ & $\text{MAE}_{\text{log}} \downarrow$ \\
    \midrule
    \midrule
    RGB only & None & 0.862 & 0.948 & 0.976 & 5.613 & 2.504 & 0.126 & 0.050 \\
    \midrule
    Sparse-to-dense~\cite{sparse2dense} & \multirow{4}{*}{Radar} & 0.876 & 0.949 & 0.974 & 5.628 & 2.374 & 0.115 & 0.047 \\
    PnP~\cite{pnp} & & 0.863 & 0.948 & 0.976 & 5.578 & 2.496 & 0.128 & 0.050 \\
    PnP + Late fusion & & 0.882 & 0.952 & 0.976 & 5.404 & 2.289 & 0.115 & 0.045 \\
    PnP + Late fusion + Robust loss & & 0.882 & 0.952 & 0.976 & 5.406 & 2.29 & 0.115 & 0.045 \\
    \cline{2-2}
    CSPN-RGB~\cite{cspn} & None & 0.864 & 0.949 & 0.976 & 5.585 & 2.478 & 0.123 & 0.049 \\
    \cline{2-2}
    CSPN-SD~\cite{cspn} & Radar & 0.867 & 0.949 & 0.976 & 5.566 & 2.457 & 0.121 & 0.048 \\
    \midrule
    \midrule
    Ours single-stage & \multirow{3}{*}{Radar} & 0.884 & 0.953 & 0.977 & 5.409 & 2.27 & 0.112 & 0.045\\
    Ours two-stage w/o smoothness & & 0.899 & 0.958 & 0.978 & 5.189 & 2.086 & 0.102 & 0.041\\
    Ours two-stage w/ smoothness & & \textbf{0.901} & \textbf{0.958} & \textbf{0.978} & \textbf{5.18} & \textbf{2.061} & \textbf{0.1} & \textbf{0.04}\\
    \bottomrule
    \end{tabular}
    \caption{Comparison to the competing methods.}
    \label{tab:exp-sota}
\end{table*}

\textbf{Input pattern experiments.} As a proof of concept, we perform experiments on a fixed model with different sparse input patterns including:
\begin{itemize}
    \item Radar depth map (\textit{Radar}): The $x_{Radar}$ created by projecting Radar point set $R$ to 2D image using perspective projection.
    \item Radar depth map filtered by LiDAR ground truth (\textit{Radar-gt-filtered}): Created by filtering outlier noises in Radar point set $R$ by comparing the depth value of the target point with its spatial neighbors in the ground truth LiDAR point set $L$. The resulting depth map is not entirely noise-free, but most of the obvious noisy measurements are removed \footnotemark.
    \item LiDAR map with Radar pattern (\textit{lidar-sampled}): The sparse depth map are created by sampling k-nearest neighbor points from LiDAR points set $L$ using Radar point locations. 
    That means the sparse depth maps will also have limited vertical field of view, but are totally noise-free because we treated LiDAR points as ground truths (k is set to 2 to keep similar point counts to the Radar depth map $x_{Radar}$).
    \item Uniformly sampled LiDAR depth map (\textit{lidar-uniform}): 
    Created by uniformly sampling the ground truth LiDAR point set $L$~\cite{sparse2dense, self-sparse2dense}. These sparse depth map have (1) no limited vertical field of view and are (2) noise-free,  which are most ideal compared with raw Radar depth map and LiDAR map with Radar pattern.
\end{itemize}
\footnotetext{Due to space limit, we didn't include the details of the filtering method.}

\renewcommand{\arraystretch}{1.5}
\begin{table*}[t!]
    \centering
    \begin{tabular}{c|c|llllll}
    \toprule 
    Methods & \makecell{Depth input\\pattern} & $\delta_{1} \uparrow$ & $\delta_{2} \uparrow$ & RMSE $\downarrow$ & MAE $\downarrow$ & REL $\downarrow$ & $\text{MAE}_{\text{log}} \downarrow$ \\
    \midrule
    \midrule
    RGB only & None & 0.874 & 0.954 & 5.44 & 2.37 & 0.12 & 0.047 \\
    \midrule
    \multicolumn{8}{c}{Daytime Experiments} \\
    \midrule
    Early fusion & \multirow{3}{*}{Radar} & 0.874 (+0\%) & 0.951 \red{(-0.31\%)} & 5.574 \red{(+2.46\%)} & 2.355 \green{(-0.63\%)} & 0.12 (+0\%) & 0.047 \red{(+0.42\%)}\\
    Ours single-stage & & 0.894 \green{(+2.24\%)} & 0.957 \green{(+0.28\%)} & 5.271 \green{(-3.1\%)} & 2.157 \green{(-8.99\%)} & 0.107 \green{(-10.92\%)} & 0.043 \green{(-9.19\%)}\\
    \makecell{Ours two-stage\\w/ smoothness} & & 0.910 \green{(+4.16\%)} & 0.962 \green{(+0.84\%)} & 5.030 \green{(-7.54\%)} & 1.941 \green{(-18.14\%)} & 0.095 \green{(-20.83\%)} & 0.038 \green{(-19.79\%)}\\
    \bottomrule
    \multicolumn{8}{c}{Nighttime Experiments} \\
    \midrule
    Early fusion & \multirow{3}{*}{Radar} & 0.795 \green{(+3.04\%)} & 0.917 \green{(+0.88\%)} & 6.723 \green{(-2.03\%)} & 3.295 \green{(-6.45\%)} & 0.159 \green{(-5.36\%)} & 0.065 \green{(-6.67\%)}\\
    Ours single-stage & & 0.814 \green{(+5.43\%)} & 0.925 \green{(+1.72\%)} & 6.402 \green{(-6.7\%)} & 3.096 \green{(-12.1\%)} & 0.147 \green{(-12.5\%)} & 0.060 \green{(-13.14\%)}\\
    \makecell{Ours two-stage\\w/ smoothness} & & 0.832 \green{(+7.79\%)} & 0.932 \green{(+2.49\%)} & 6.29 \green{(-8.34\%)} & 2.933 \green{(-16.72\%)} & 0.135 \green{(-19.64\%)} & 0.056 \green{(-19.47\%)}\\
    \bottomrule
    \end{tabular}
    \caption{Daytime and nighttime: Improvements compared with RGB only model (1st row). (Top part) daytime. (Bottom part) nighttime. Percentage in the bracket denotes the relative improvements compared with the RGB only model (1st row). \green{Green} stands for improvements and \red{Red} stands for degradation.}
    \label{tab:exp-daynight}
\end{table*}

In \cref{tab:exp-pattern}, we list the performance using different depth input patterns. 
Note that here we fix the model to the standard encoder-decoder with early fusion. 
Apparently, the last row (LiDAR uniform) is the best because there are no noisy measurements and limited vertical field of view issue. 
However, by comparing LiDAR with Radar pattern (4th row) and Radar filtered by LiDAR ground truth (3rd-row) to the Radar (2nd row), we can see that certain amount of improvements can still be made in spite of the limited vertical field of view (\textit{lidar-sampled}) and inconsistent measurements (\textit{Radar-gt-filtered}) using the simple early fusion model, which means that removing noisy measurements is indispensable in order to have improvements under the field of view and sparseness limitations. 
This is the main motivation behind our two-stage design (\cref{subsec:cnn}) and noise filtering module (\cref{subsec:noise_filter}).


\begin{figure*}[t]
\centering
\includegraphics[width=1.0\textwidth]{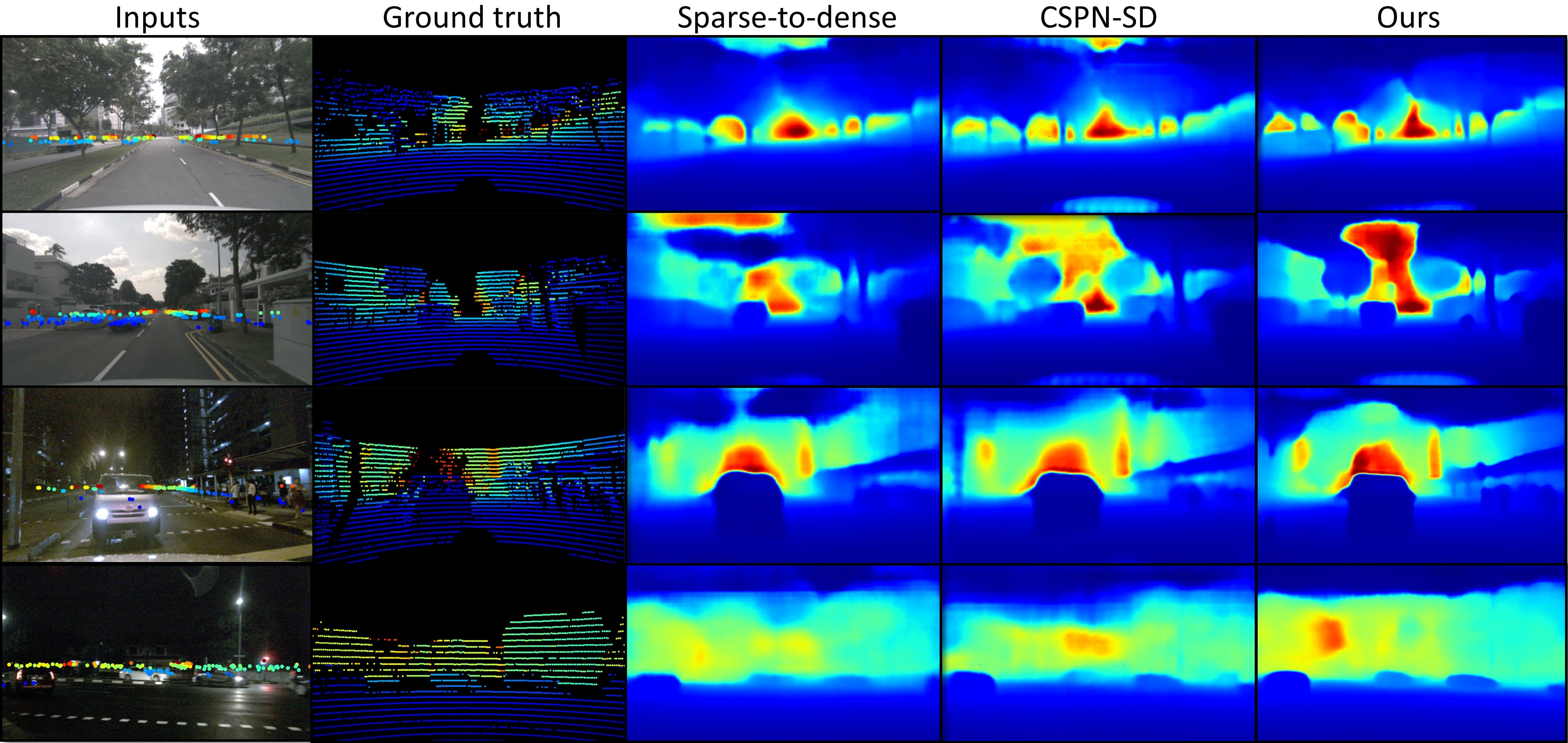} 
\caption{Qualitative results. We overlay the radar points on RGB images. From the figure, we can see that the predictions from our method have better quality in both daytime and nighttime. Here ours stands for our full model (Ours two-stagew/ smoothness)}
\vspace{-1mm}
\label{fig:qual_result}
\end{figure*}

\subsection{Comparison to competing methods}
\label{subsec:exp-sota}
To demonstrate the effectiveness of our method, we compare the performance of our model with existing RGB + LiDAR fusion methods~\cite{sparse2dense, cspn, pnp} on our RGB + Radar fusion task.
Below we provide the details about the completing methods used in the comparison:

\textbf{Sparse-to-dense~\cite{sparse2dense}.}\footnotemark We use the same encoder and decoder design as the backbone of our model. 

\footnotetext{Official implementation: \href{https://github.com/fangchangma/sparse-to-dense}{https://github.com/fangchangma/sparse-to-dense}}


\textbf{PnP~\cite{pnp}.}\footnotemark Following the suggestion from the author, we use $\alpha=0.01$ and perform 5 refinement iterations. 
Since the method takes the given sparse depth maps $x_{Radar}$ directly as ground truths and perform the refinement, the refinement results on the points with inconsistent or noisy measurements will actually damage the output predictions instead.
Apart from combining PnP with the RGB only model (the version mentioned in their paper), we further try to combine it with our late fusion model (PnP + Late fusion) and use different loss function that is robust to outliers~\cite{robust-loss} (PnP + Late fusion + Robust loss). 
However, their results are similar to the late fusion model (Ours single-stage).

\footnotetext{Official implementation: \href{https://github.com/zswang666/PnP-Depth}{https://github.com/zswang666/PnP-Depth}}

\textbf{CSPN~\cite{cspn}.}\footnotemark
To have a fair comparison, we adapted the implementation from the author to use the same Resnet18~\cite{resnet} backbone as our methods and Sparse-to-dense~\cite{sparse2dense}. 
Following the official implementation, we set the number of refinement iterations to 20. 
In \cref{tab:exp-sota}, CSPN-RGB stands for the RGB only version of CSPN~\cite{cspn}, and CSPN-SD stands for the CSPN~\cite{cspn} model taking sparse Radar depth maps $x_{Radar}$ into account in the refinement stage (in the original work, the refinement stage takes ground truth LiDAR depth maps as inputs). 
\footnotetext{Official implementation: \href{https://github.com/XinJCheng/CSPN}{https://github.com/XinJCheng/CSPN}}

The results are presented in \cref{tab:exp-sota}.
From the table, we can see that if taking raw Radar depth maps as inputs, most of the method cannot yield improvements over the RGB-only baseline.  On the contrary, our two-stage model can effectively extract useful information from the noisy measurements and improve the final depth predictions. This clearly show that the denoising stage is crucial for effectively using the noisy Radar data. 

In addition to the quantitative results, we also provide some qualitative results in \cref{fig:qual_result}. We can see that the predictions from our method retain more details than CSPN~\cite{cspn} and Sparse-to-dense~\cite{sparse2dense} both in daytime and nighttime scenarios.

\subsection{Ablation study}
\label{subsec:exp-ablation}
To verify the effectiveness of each component of our full model, we also include the ablation study in the bottom part of \cref{tab:exp-sota}. Here are the details of our model's variants used in ablation study:
\begin{itemize}
    \item \textbf{Ours single-stage:} The late fusion model introduced in \cref{subsec:cnn}. No filtering operation is performed to remove noisy measurements.
    \item \textbf{Ours two-stage w/o smoothness:} The whole two stage model with noise filtering module (\cref{subsec:noise_filter}), but without smoothness loss in stage1, and $w1$ and $w2$ in \cref{eq:total} are set to 0 (1:1 on losses of stage1 and stage2).
    \item \textbf{Ours two-stage w/ smoothness:} The final version of our model with all the components mentioned in \cref{sec:method}.
\end{itemize}
We can see that the performance drops with each component disabled, and our full model has the best performance.

\subsection{Day-Night experiment}
\label{subsec:exp-day-night}

To further demonstrate the benefits from combining Radar measurements with RGB images, we analyze the improvements on different lighting conditions (daytime and nighttime views). 
By comparing the relative improvements in \cref{tab:exp-daynight}, we can first find that the performance of all models in daytime are much better than in nighttime as we expected. 
Combining information from Radar measurements, the relative improvements in nighttime scenario are generally better than daytime ones. 
Even the early fusion model has improvements on all the metrics.
This further illustrates the increased robustness from Radar measurements in nighttime scenarios where the RGB only model does not perform well.


\section{CONCLUSIONS}
We have proposed a novel method for dense depth estimation by fusing monocular images and sparse Radar data. We have shown that why existing methods developed for LiDAR data do not work for our task and motivated our method carefully by our experimental observations. Extending the work to videos constitutes our future work.

\noindent
\textbf{Acknowledgement}:
The work is supported by Toyota via the research project TRACE-Zurich. 

\bibliographystyle{IEEEtran.bst}
\bibliography{references}

\begin{thebibliography}{10}
\providecommand{\url}[1]{#1}
\csname url@rmstyle\endcsname
\providecommand{\newblock}{\relax}
\providecommand{\bibinfo}[2]{#2}
\providecommand\BIBentrySTDinterwordspacing{\spaceskip=0pt\relax}
\providecommand\BIBentryALTinterwordstretchfactor{4}
\providecommand\BIBentryALTinterwordspacing{\spaceskip=\fontdimen2\font plus
\BIBentryALTinterwordstretchfactor\fontdimen3\font minus
  \fontdimen4\font\relax}
\providecommand\BIBforeignlanguage[2]{{%
\expandafter\ifx\csname l@#1\endcsname\relax
\typeout{** WARNING: IEEEtran.bst: No hyphenation pattern has been}%
\typeout{** loaded for the language `#1'. Using the pattern for}%
\typeout{** the default language instead.}%
\else
\language=\csname l@#1\endcsname
\fi
#2}}

\bibitem{nyudepthv2}
P.~K. Nathan~Silberman, Derek~Hoiem and R.~Fergus, ``Indoor segmentation and
  support inference from rgbd images,'' in \emph{European Conference on
  Computer Vision (ECCV)}, 2012.

\bibitem{dorn}
H.~Fu, M.~Gong, C.~Wang, K.~Batmanghelich, and D.~Tao, ``Deep ordinal
  regression network for monocular depth estimation,'' in \emph{IEEE Conference
  on Computer Vision and Pattern Recognition (CVPR)}, 2018.

\bibitem{depth-cnn1}
R.~Garg, V.~K. BG, G.~Carneiro, and I.~Reid, ``Unsupervised cnn for single view
  depth estimation: Geometry to the rescue,'' in \emph{European Conference on
  Computer Vision (ECCV)}, 2016.

\bibitem{fcrn}
I.~Laina, C.~Rupprecht, V.~Belagiannis, F.~Tombari, and N.~Navab, ``Deeper
  depth prediction with fully convolutional residual networks,'' in
  \emph{International Conference on 3D Vision (3DV)}, 2016.

\bibitem{reccurent_depth_2020}
P.~Vaishakh, W.~Van~Gansbeke, D.~Dai, and L.~Van~Gool, ``Don't forget the past:
  Recurrent depth estimation from monocular video,'' in \emph{IROS}, 2020.

\bibitem{deeplidar}
J.~Qiu, Z.~Cui, Y.~Zhang, X.~Zhang, S.~Liu, B.~Zeng, and M.~Pollefeys,
  ``Deeplidar: Deep surface normal guided depth prediction for outdoor scene
  from sparse lidar data and single color image,'' in \emph{IEEE Conference on
  Computer Vision and Pattern Recognition (CVPR)}, 2019.

\bibitem{sparse2dense}
F.~Mal and S.~Karaman, ``Sparse-to-dense: Depth prediction from sparse depth
  samples and a single image,'' in \emph{IEEE International Conference on
  Robotics and Automation (ICRA)}, 2018.

\bibitem{uber-conti-2}
M.~Liang, B.~Yang, S.~Wang, and R.~Urtasun, ``Deep continuous fusion for
  multi-sensor 3d object detection,'' in \emph{European Conference on Computer
  Vision (ECCV)}, 2018.

\bibitem{uber-conti-3}
M.~Liang, B.~Yang, Y.~Chen, R.~Hu, and R.~Urtasun, ``Multi-task multi-sensor
  fusion for 3d object detection,'' in \emph{IEEE Conference on Computer Vision
  and Pattern Recognition (CVPR)}, 2019.

\bibitem{kitti}
A.~Geiger, P.~Lenz, and R.~Urtasun, ``Are we ready for autonomous driving? the
  kitti vision benchmark suite,'' in \emph{IEEE Conference on Computer Vision
  and Pattern Recognition (CVPR)}, 2012.

\bibitem{nuscenes2019}
H.~Caesar, V.~Bankiti, A.~H. Lang, S.~Vora, V.~E. Liong, Q.~Xu, A.~Krishnan,
  Y.~Pan, G.~Baldan, and O.~Beijbom, ``nuscenes: A multimodal dataset for
  autonomous driving,'' in \emph{IEEE Conference on Computer Vision and Pattern
  Recognition (CVPR)}, 2020.

\bibitem{early_stereo1}
D.~Scharstein and R.~Szeliski, ``A taxonomy and evaluation of dense two-frame
  stereo correspondence algorithms,'' \emph{International Journal of Computer
  Vision (IJCV)}, 2002.

\bibitem{early_stereo2}
J.~Flynn, I.~Neulander, J.~Philbin, and N.~Snavely, ``Deepstereo: Learning to
  predict new views from the world's imagery,'' in \emph{IEEE Conference on
  Computer Vision and Pattern Recognition (CVPR)}, 2016.

\bibitem{make3d}
A.~Saxena, M.~Sun, and A.~Y. Ng, ``Make3d: Learning 3d scene structure from a
  single still image,'' \emph{IEEE Transactions on Pattern Analysis and Machine
  Intelligence (TPAMI)}, 2008.

\bibitem{early_mono1}
L.~Ladicky, J.~Shi, and M.~Pollefeys, ``Pulling things out of perspective,'' in
  \emph{IEEE Conference on Computer Vision and Pattern Recognition (CVPR)},
  2014.

\bibitem{early_mono2}
D.~Hoiem, A.~A. Efros, and M.~Hebert, ``Recovering surface layout from an
  image,'' \emph{International Journal of Computer Vision (IJCV)}, 2007.

\bibitem{vgg}
K.~Simonyan and A.~Zisserman, ``Very deep convolutional networks for
  large-scale image recognition,'' in \emph{International Conference on
  Learning Representations (ICLR)}, 2015.

\bibitem{resnet}
K.~He, X.~Zhang, S.~Ren, and J.~Sun, ``Deep residual learning for image
  recognition,'' in \emph{IEEE Conference on Computer Vision and Pattern
  Recognition (CVPR)}, 2016.

\bibitem{fcn}
J.~Long, E.~Shelhamer, and T.~Darrell, ``Fully convolutional networks for
  semantic segmentation,'' in \emph{IEEE Conference on Computer Vision and
  Pattern Recognition (CVPR)}, 2015.

\bibitem{mask-rcnn}
K.~He, G.~Gkioxari, P.~Doll{\'a}r, and R.~Girshick, ``Mask r-cnn,'' in
  \emph{IEEE International Conference on Computer Vision (ICCV)}, 2017.

\bibitem{cityscapes}
M.~Cordts, M.~Omran, S.~Ramos, T.~Rehfeld, M.~Enzweiler, R.~Benenson,
  U.~Franke, S.~Roth, and B.~Schiele, ``The cityscapes dataset for semantic
  urban scene understanding,'' in \emph{IEEE Conference on Computer Vision and
  Pattern Recognition (CVPR)}, 2016.

\bibitem{mscoco}
T.-Y. Lin, M.~Maire, S.~Belongie, J.~Hays, P.~Perona, D.~Ramanan,
  P.~Doll{\'a}r, and C.~L. Zitnick, ``Microsoft {COCO}: Common objects in
  context,'' in \emph{European Conference on Computer Vision (ECCV)}, 2014.

\bibitem{eigen}
D.~Eigen and R.~Fergus, ``Predicting depth, surface normals and semantic labels
  with a common multi-scale convolutional architecture,'' in \emph{IEEE
  International Conference on Computer Vision (ICCV)}, 2015.

\bibitem{depth-cnn2}
J.~Li, R.~Klein, and A.~Yao, ``A two-streamed network for estimating
  fine-scaled depth maps from single rgb images,'' in \emph{IEEE International
  Conference on Computer Vision (ICCV)}, 2017.

\bibitem{depth-skip}
J.~Xie, R.~Girshick, and A.~Farhadi, ``Deep3d: Fully automatic 2d-to-3d video
  conversion with deep convolutional neural networks,'' in \emph{European
  Conference on Computer Vision (ECCV)}, 2016.

\bibitem{sfmlearner}
T.~Zhou, M.~Brown, N.~Snavely, and D.~G. Lowe, ``Unsupervised learning of depth
  and ego-motion from video,'' in \emph{IEEE Conference on Computer Vision and
  Pattern Recognition (CVPR)}, 2017.

\bibitem{unsupervised-photo}
R.~Garg, V.~K. BG, G.~Carneiro, and I.~Reid, ``Unsupervised cnn for single view
  depth estimation: Geometry to the rescue,'' in \emph{European Conference on
  Computer Vision (ECCV)}, 2016.

\bibitem{semi-smooth}
Y.~Kuznietsov, J.~Stuckler, and B.~Leibe, ``Semi-supervised deep learning for
  monocular depth map prediction,'' in \emph{IEEE Conference on Computer Vision
  and Pattern Recognition (CVPR)}, 2017.

\bibitem{monodepth2}
C.~Godard, O.~{Mac Aodha}, M.~Firman, and G.~J. Brostow, ``Digging into
  self-supervised monocular depth prediction,'' in \emph{IEEE International
  Conference on Computer Vision (ICCV)}, 2019.

\bibitem{edge-aware-smoothness}
C.~Godard, O.~Mac~Aodha, and G.~J. Brostow, ``Unsupervised monocular depth
  estimation with left-right consistency,'' in \emph{IEEE Conference on
  Computer Vision and Pattern Recognition (CVPR)}, 2017.

\bibitem{completion1}
M.~Jaritz, R.~De~Charette, E.~Wirbel, X.~Perrotton, and F.~Nashashibi, ``Sparse
  and dense data with cnns: Depth completion and semantic segmentation,'' in
  \emph{IEEE International Conference on 3D Vision (3DV)}, 2018.

\bibitem{self-sparse2dense}
F.~Ma, G.~V. Cavalheiro, and S.~Karaman, ``Self-supervised sparse-to-dense:
  Self-supervised depth completion from lidar and monocular camera,'' in
  \emph{IEEE International Conference on Robotics and Automation (ICRA)}, 2019.

\bibitem{pnp}
T.-H. Wang, F.-E. Wang, J.-T. Lin, Y.-H. Tsai, W.-C. Chiu, and M.~Sun,
  ``Plug-and-play: Improve depth estimation via sparse data propagation,'' in
  \emph{International Conference on Robotics and Automation (ICRA)}, 2018.

\bibitem{adversarial}
A.~Kurakin, I.~Goodfellow, and S.~Bengio, ``Adversarial examples in the
  physical world,'' \emph{arXiv preprint arXiv:1607.02533}, 2016.

\bibitem{cspn}
X.~Cheng, P.~Wang, and R.~Yang, ``Depth estimation via affinity learned with
  convolutional spatial propagation network,'' in \emph{European Conference on
  Computer Vision (ECCV)}, 2018.

\bibitem{Chadwick2019DistantVD}
S.~Chadwick, W.~Maddern, and P.~Newman, ``Distant vehicle detection using radar
  and vision,'' \emph{International Conference on Robotics and Automation
  (ICRA)}, 2019.

\bibitem{nobis19crfnet}
F.~Nobis, M.~Geisslinger, M.~Weber, J.~Betz, and M.~Lienkamp, ``A deep
  learning-based radar and camera sensor fusion architecture for object
  detection,'' in \emph{Sensor Data Fusion: Trends, Solutions, Applications
  (SDF)}, 2019.

\bibitem{john2019so}
V.~John, M.~Nithilan, S.~Mita, H.~Tehrani, R.~Sudheesh, and P.~Lalu, in
  \emph{Pacific-Rim Symposium on Image and Video Technology}, 2019.

\bibitem{radar:dataset:19}
M.~{Meyer} and G.~{Kuschk}, ``Automotive radar dataset for deep learning based
  3d object detection,'' in \emph{European Radar Conference (EuRAD)}, 2019.

\bibitem{semseg:radar:points:cloud:18}
O.~{Schumann}, M.~{Hahn}, J.~{Dickmann}, and C.~{Wöhler}, ``Semantic
  segmentation on radar point clouds,'' in \emph{International Conference on
  Information Fusion (FUSION)}, 2018.

\bibitem{soundperception20}
A.~B. Vasudevan, D.~Dai, , and L.~{Van Gool}, ``Semantic object prediction and
  spatial sound super-resolution with binaural sounds,'' in \emph{ECCV}, 2020.

\bibitem{multi_uncertainty}
A.~Kendall, Y.~Gal, and R.~Cipolla, ``Multi-task learning using uncertainty to
  weigh losses for scene geometry and semantics,'' in \emph{IEEE Conference on
  Computer Vision and Pattern Recognition (CVPR)}, 2018.

\bibitem{pytorch}
A.~Paszke, S.~Gross, F.~Massa, A.~Lerer, J.~Bradbury, G.~Chanan, T.~Killeen,
  Z.~Lin, N.~Gimelshein, L.~Antiga, \emph{et~al.}, ``Pytorch: An imperative
  style, high-performance deep learning library,'' in \emph{Advances in Neural
  Information Processing Systems (NeurIPS)}, 2019.

\bibitem{imagenet}
J.~Deng, W.~Dong, R.~Socher, L.-J. Li, K.~Li, and L.~Fei-Fei, ``Imagenet: A
  large-scale hierarchical image database,'' in \emph{IEEE Conference on
  Computer Vision and Pattern Recognition (CVPR)}, 2009.

\bibitem{robust-loss}
J.~T. Barron, ``A general and adaptive robust loss function,'' in \emph{IEEE
  Conference on Computer Vision and Pattern Recognition (CVPR)}, 2019.

\end{thebibliography}

\normalsize

\end{document}